\documentclass[10pt,twocolumn,letterpaper]{article}
\usepackage{graphicx, wrapfig, lipsum}
\usepackage{comment}
\usepackage{amsmath,amssymb} 
\usepackage{breqn}
\newcommand{\R}{\mathbb{R}}
\newcommand{\EE}{\mathbb{E}}

\usepackage{color}
\usepackage{cite}
\usepackage{booktabs}

\newcommand{\btheta}{\boldsymbol{\theta}}
\newcommand{\bphi}{\boldsymbol{\phi}}

\newcommand{\bx}{\mathbf{x}}

\usepackage{wacv}
\usepackage{times}
\usepackage{epsfig}
\usepackage{graphicx}
\usepackage{amsmath}
\usepackage{amssymb}

\def\wacvPaperID{1002} 

\wacvfinalcopy 
\ifwacvfinal
\def\assignedStartPage{9876} 
\fi

\ifwacvfinal
\usepackage[breaklinks=true,bookmarks=false]{hyperref}
\else
\usepackage[pagebackref=true,breaklinks=true,colorlinks,bookmarks=false]{hyperref}
\fi

\ifwacvfinal
\else
\fi

\begin{document}

\title{Optimistic Agent: Accurate Graph-Based Value Estimation for More Successful Visual Navigation}

\author{Mahdi Kazemi Moghaddam, Qi Wu, Ehsan Abbasnejad and Javen Shi\\
{\small Australian Institute for Machine Learning}\\
{\small The University of Adelaide}\\
 {\tt\small mahdi.kazemimoghaddam, qi.wu01, ehsan.abbasnejad and javen.shi@adelaide.edu.au}

}

\maketitle


\begin{abstract}
We humans can impeccably search for a target object, given its name only, even in an unseen environment. We argue that this ability is largely due to three main reasons: the incorporation of \emph{prior knowledge} (or experience), the adaptation of it to the new environment using the observed \emph{visual cues} and most importantly \emph{optimistically} searching without giving up early. This is currently missing in the state-of-the-art visual navigation methods based on Reinforcement Learning (RL). In this paper, we propose to use externally learned prior knowledge of the relative object locations and integrate it into our model by constructing a neural graph. In order to efficiently incorporate the graph without increasing the state-space complexity, we propose Graph-based Value Estimation (GVE) module. GVE provides a more accurate baseline for estimating the \textit{Advantage} function in actor-critic RL algorithm. This results in reduced value estimation error and, consequently, convergence to a more optimal policy. Through empirical studies, we show that our agent, dubbed as the \emph{optimistic} agent, has a more realistic estimate of the state value during a navigation episode which leads to a higher success rate. Our extensive ablation studies show the efficacy of our simple method which achieves the state-of-the-art results measured by the conventional visual navigation metrics, e.g. Success Rate (SR) and Success weighted by Path Length (SPL), in AI2THOR environment.
\end{abstract}


\section{Introduction}

Human beings are capable of finding a given object in an unexplored environment, \textit{e.g.} a room in a new house, given just a short natural language command, \textit{e.g.} name of the object. We primarily rely on our prior knowledge of the way different objects are co-located in a specific room in addition to the new sensory input information we receive. For example, we know in any bathroom, soap bottle should be located near the basin, thus observing one (assuming limited field of view and/or longer distance) helps us find the other more easily (\textit{e.g.} by providing clues). Our belief, however, needs to be adjusted upon receiving new observations in a previously unseen environment. For example, the soap bottle might be misplaced or it might be in a new unseen shape or color. Finally, we do not normally give up on searching as soon as we fail to find the target in the first attempt. In other words, we optimistically continue our search as long as we receive indications from the environment that the target is reachable. It is desirable for every navigation agent to be able to utilise prior knowledge, being prepared for adapting the beliefs to a new unseen environment and being robust to early capitulation.

\begin{figure}[t]
\includegraphics[width=8cm]{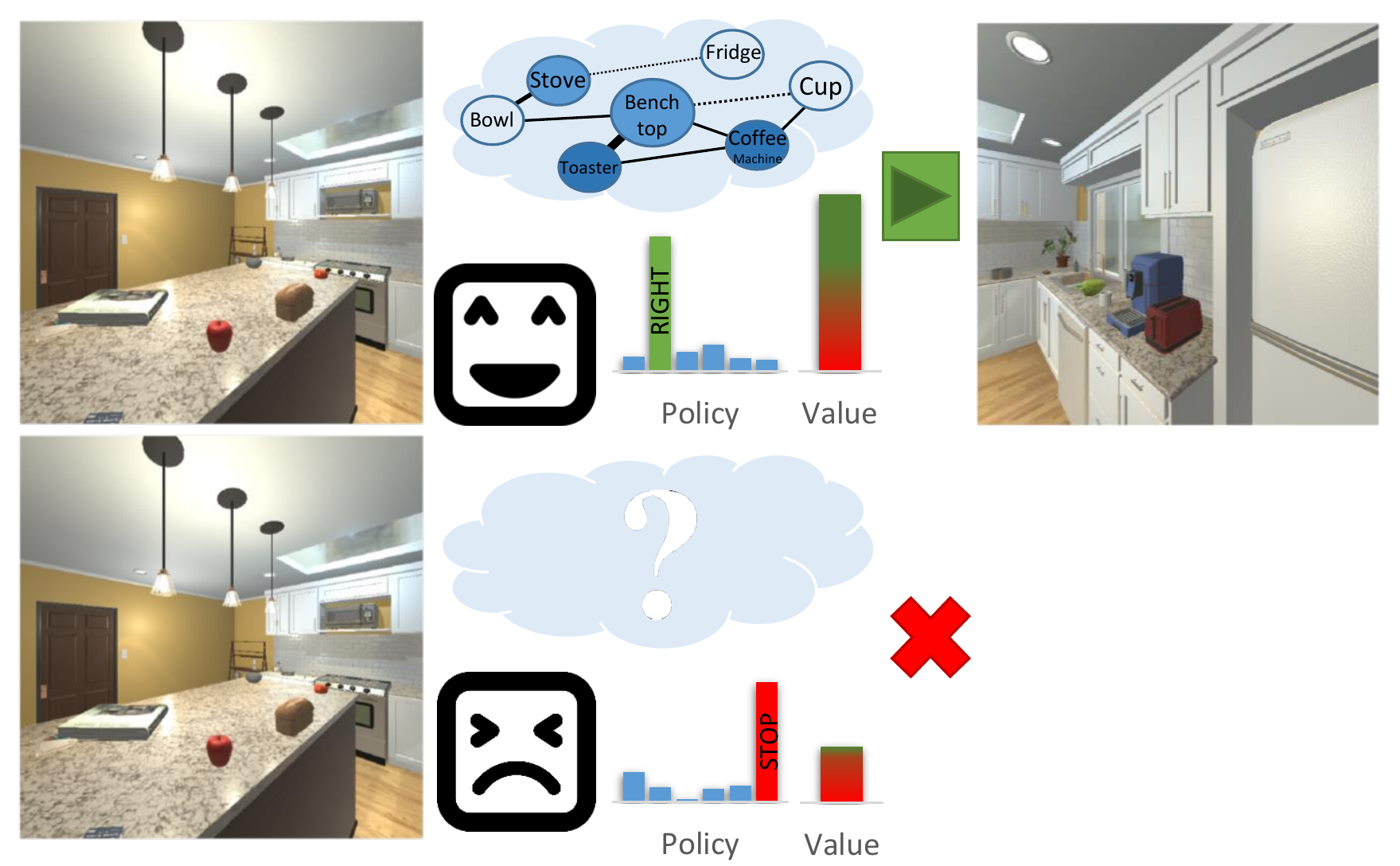}
\centering
\caption{Our optimistic agent estimates the state value more accurately with the help of a graph representing the relationship between object locations. The more accurate value estimation helps with learning a more optimal policy that is robust to early capitulation (\textit{e.g.} stopping before reaching the target object as a result of under-estimating the state value).}
\label{motivation}
\vspace{-5pt}
\end{figure}


To endow the navigation agents with the first capability, recent works \cite{scene_priors, kg_class, kg_situation} have used Graph Neural Networks (GNNs) to encode the object-object relationships in a pre-trained model, in the form of a knowledge graph. Their method, however, as we show in our experiments, is not efficient and cannot scale to more complex models. The major drawback of those methods is increasing the complexity of the state-space. This is crucial to avoid in sparse reward RL set-up since it exacerbates the challenges around credit assignment and impedes the performance improvement. On the other hand, to enable the agents with the second skill (e.g. adaptation), there is an existing work \cite{savn} which incorporates meta-learning to allow the agent to quickly adapt to a new environment, using a few gradient-based weight updates. While the mentioned work shows relative success in handling the train-test distribution shift, its efficiency in adapting prior knowledge to unobserved scenes has remained unexplored. This task, as will be shown in the ablation studies, is non-trivial in RL framework. The reason is as we increase the model's complexity, efficient training becomes more challenging with conventional RL objective functions.

Finally, the last mentioned desirable capability for a navigation agent to have is a realistic (read it as optimistic) estimation of it's expected future reward. Revisiting the bathroom example again, an optimistic agent will continue to search for a soap bottle that is not found in its immediate expected location, e.g. near the basin. An unrealistic agent, however, will give up as soon as it fails to find the soap bottle near the basin without further search. In actor-critic RL a more realistic estimation of the value function will lead to a more optimal policy that is more robust to early capitulation.

To that end, we propose our Graph-based Value Estimation (GVE) module to efficiently benefit from the prior knowledge for more accurate value function estimation. The prior knowledge presents the agent with general rules about the semantic relationships of the objects and their relative location in an indoor environment. Instead of using the graph as the input to the state representation encoder, which uncontrollably increases the effective size of the state space, we use the graph to estimate the potential return \footnote{Defined as the expected sum of discounted future rewards.\ref{setup}} from the current state. This way the graph is used to better guide the policy training rather than being directly used for action taking. We present, in our experiments, that simply conditioning the policy on the prior knowledge does not lead to a more optimal policy with better performance.

Intuitively, observing a correlated object to the target that the agent is searching for has a strong indication of the final reward that can be gained, following the current policy. For instance, when searching for a book in a room, observing a desk or shelves from the distance has an indication of finding the book, hence the current state has a higher value. The association of these observations through unstructured trial and error in RL without incorporating the prior knowledge is not simply achievable; specifically, when objects are small the extracted visual features might not contain much information to guide the agent towards them. 

Our GVE results in a more optimally trained critic that can associate the visual observations with its prior knowledge to better estimate the state value. This is essential in our framework where we use a variant of actor-critic RL algorithm (\textit{e.g.} Asynchronous Advantage Actor-Critic; A3C \cite{a3c}, where the critic's role is to identify the actions leading to a specific reward using the temporal difference of the state values. We empirically show that GVE improves the value estimation accuracy. It is shown that the high accuracy of value estimation is essential for an optimal policy training \cite{reinforce_1992}, which is further confirmed in the performance of our proposed method.



In overall, we propose a simple, principled and modular approach that can be employed in conjunction with other approaches for visual navigation that also use an actor-critic RL framework. Our main contributions are:
\begin{itemize}
    
    \item We introduce a simple, yet effective method that can endow an actor-critic navigation agent by improved integration of prior knowledge for navigation;
    
    \item We propose the GVE module which reduces the value estimation error in A3C RL algorithm. We empirically show the higher accuracy of our estimates which leads to a more optimal policy.
    
    \item Finally, we provide the new state-of-the-art performance results on the object-target visual navigation task in AI2THOR environment. This is achieved in the more challenging scenario of navigation using target object name only.
\end{itemize}


\section{Related Work} \label{rel_work}
\subsection{Visual Navigation}
End-to-end visual navigation has recently gained attention \cite{vis_nav_eval,learn_to_nav, citi_nav, nav_out, cmp, thor_target_driven} and different tasks have been introduced. Some tasks consider inputting images of the target~\cite{thor_target_driven, cmp} while others consider the more challenging task of natural language instructions \cite{vis_nav_eval, scene_priors, look_before_leap, savn, r2r}. In the latter the agent has to ground language instructions on observations while performing planning and navigation. The main approaches can be divided into supervised (\textit{e.g.} Imitation Learning) \cite{r2r} and unsupervised (RL) \cite{learn_to_nav} methods.

The RL-based navigation has gained extra momentum in the past few years owing to the availability of high-quality 3D simulators \cite{r2r, habitat, thor_env, gibson} as well as more efficient deep RL algorithms \cite{humanlevelcontrol, contcontrol}. AI2THOR \cite{thor_env} is a photo-realistic simulator that unlike some other simulators \cite{r2r} enables continuous movements of the agent. This feature, while making the navigation task more challenging is closer to real-world scenarios, hence more valuable. In this environment, it is very likely for a sub-optimal agent to stand in front of a blocked path, e.g. an obstacle, and continuously perform a failing action, e.g. move forward, until the maximum step limit is exhausted. 

\subsection{Prior Knowledge}
Adding some form of prior to machine learning models has fast-tracked the progress in many different computer vision tasks \cite{scene_graph_gen, prior_rl, prior_nav_mid_level, prior_det_seg}. The prior can be in the form of pre-trained weights of the network for object detection or segmentation tasks \cite{prior_nav_mid_level} or in the from of Graph Neural Networks (GNNs) representing common sense knowledge \cite{scene_graph_gen}. 
In RL framework, GNNs have been used to improve navigation models by representing topological environment maps for localisation \cite{graph_topo} or helping with more efficient exploration \cite{graph_explore}. Generating and incorporating scene graphs \cite{scene_graph_gen} is also closely related to our problem. However, here we construct our knowledge graph externally and refine the node relationships without explicitly detecting the objects in the observation. This also separates our approach from \cite{avd_object_nav} and \cite{anu_nav}, where an off-the-shelf object detector is used. These methods, while improving the performance, have been explored before and can be plugged into other approaches for further improvements, including ours. The other difference of the concurrent work of Du \etal \cite{anu_nav} to ours is the use of Imitation Learning (IL), which further makes it incomparable since here we only use sparse rewards and not ground-truth actions for training. 

A similar work to ours is proposed by Vijay \etal~\cite{prior_graph_edge} where the prior knowledge is injected into RL for navigation. The authors propose to learn various edge features encoded as one-hot vectors. The main difference, however, is that they address the navigation in a 2D fully-observable environment while in our case we deal with near-real-world task of navigation in 3D environment.

In \cite{scene_priors} the authors use a knowledge graph to learn to find unseen objects by learning the correlations to the seen targets during the training. One of the deficiencies of their method is that it effectively increases the state space size. This is specifically important in RL framework where increasing the model's complexity can impede improving the performance, as we show in this work.

\subsection{Value Function Estimation in Actor-Critic RL}
One major drawback of model-free RL methods is the high sample complexity. This is specifically worsened in our 3D environment set-up where the agent has to learn the visual associations and ground the language instruction while learning the optimal policy using only sparse rewards. In order to mitigate this in actor-critic RL, some of the parameters of the policy are shared with the value estimation module which provides a stronger learning signal for the shared parameters. An accurate and stable estimate of the value, however, is necessary to reach this objective \cite{reinforce, gae}.

In \cite{duelling_net} the authors separate the parameters of the advantage and value functions to increase the accuracy of action-value function estimation. Inspired by that, we propose a novel value estimation module for visual navigation. This module can estimate the value of each state more accurately incorporating the object relationships encoded using a graph neural network.


\section{Our Method}\label{method}
In this section, we first describe the task in details and then discuss our proposed method.
\subsection{Preliminaries}\label{setup}

The AI2THOR \cite{thor_env} task is to navigate to a target object given its name, as a word embedding vector. The navigation episode starts when the agent is spawned in a uniformly random location and orientation in one of the four rooms, \textit{e.g.} kitchen, living room, bedroom or bathroom. The room is also sampled from a uniform distribution. Then a randomly sampled target object, from among visible objects in the scene, is presented to the agent in plain language, \textit{e.g.} "fridge" or "soap", for example. The only accessible observation to the agent is its egocentric RGB image at each time step. The agent takes actions sampled from its policy based on the observation at each time step to locate the target object. An episode ends if either the agent stops within a defined distance of an instance of the target object or the maximum number of permitted actions is exhausted.
We define the problem as a Partially-Observable Markov Decision Problem (POMDP) of $\{X, A, r, \gamma\}$. Here $\{X\}$ is the state space, the action space is $A$, $r$ is the immediate reward at each step and $\gamma$ is the reward discount factor. A trajectory of $(x_0, a_0, r_0, x_1, a_1, r_1 ...)$ is generated by taking action $a_t$ at time $t$ and observing the states according to the unknown dynamics of the environment $x_{t+1} \sim P (x_{t+1} | x_t, a_t) $. A reward $r_t = r(x_t, a_t, x_{t+1})$ is received from the environment at each time stop.

In this setup, the agent is trained to maximise the accumulated expected reward, $\EE_{\tau \sim \pi} [\sum_{t = i}^{T} \gamma^t r_t]$, where $\tau$ is a trajectory  sampled from the agent's policy $\pi$. The policy is approximated by a neural network, which receives the state $x_t$ and a target object name embedding $Z$. 

We employ A3C~\cite{a3c} as one of the most effective actor-critic RL algorithms to train our policy. We focus on actor-critic methods because they are  popular in visual navigation due to their efficiency, ease of use and robustness. In addition, these methods bridge between gradient-based and value iteration approaches inheriting their benefits.
We hypothesise that our proposed method can be applied to other actor-critic methods with minimal modifications. In our approach, the model is parameterised with $=\{\btheta,\btheta_{\pi},\btheta_v \}$; 
$\btheta$ is the set of parameters of the backbone conditional state embedding network, $\btheta_{\pi}$ is the set of parameters for the policy sub-network, \textit{a.k.a} actor, and $\btheta_v$ is the set of value sub-network's parameters, \textit{a.k.a} critic. We use a CNN to encode the observations and Glove~\cite{glove} to encode the word embeddings for the target. More details on the network architecture is presented in methods and implementation details.

Conventionally, the actor and critic share the network parameters. However, they are \emph{not} tasked with the same objective: while the actor needs a minimum representation to capture the environment to take appropriate actions at each time-step, the critic needs a sufficiently holistic representation to estimate how likely it could be for the agent to achieve its goal. The critic does not need every detail of the scene and requires a global representation as opposed to that of the policy. Through the feedback the critic provides, the actor improves which is an implicit guide for the policy.
With this intuition, in our proposed approach we augment the critic's sub-network with our Graph-based Value Function Estimation (GVE) module that considers and updates a knowledge graph of the object relations. GVE leads to a more accurate value estimation which then reduces the variance of the gradient samples~\cite{reinforce_1992}. This is because the policy is updated at each time step using gradient descent to minimise the following loss functions: 
\begin{equation}
    \begin{split}
        L_{\pi}(a_t | \bx_t)  \approx & -\log\pi (a_t|\bx_t ; \btheta, \btheta_{\pi})(r_t(a_t | \bx_t) + \\&\gamma \hat{V}^{\pi}(\bx_{t+1}) - \hat{V}^{\pi}(\bx_t)) + \beta_H  H_t(\pi)\\
        L_{V}(x_t) =& \frac{1}{2}( \hat{V}^{\pi}(x_t) - V(x_t))^2
    \end{split}
\end{equation}
where $L_{\pi}$ is the policy function loss and $L_{V}$ is the value function loss. Here, $\hat{V}(x_t)$ is our method's estimate of the ground truth value $V(x_t)$ at state $x_t$, $H_t$ is the entropy of the policy at time step $t$ and $\beta_H$ is hyper-parameter to encourage exploration. It is shown in~\cite{reinforce_1992} that a more accurate estimation of the value functions leads to a lower variance of the gradients which encourages a more stable training and consequently  finding a better policy.

\begin{figure*}
\centering
\includegraphics[width=0.8\textwidth]{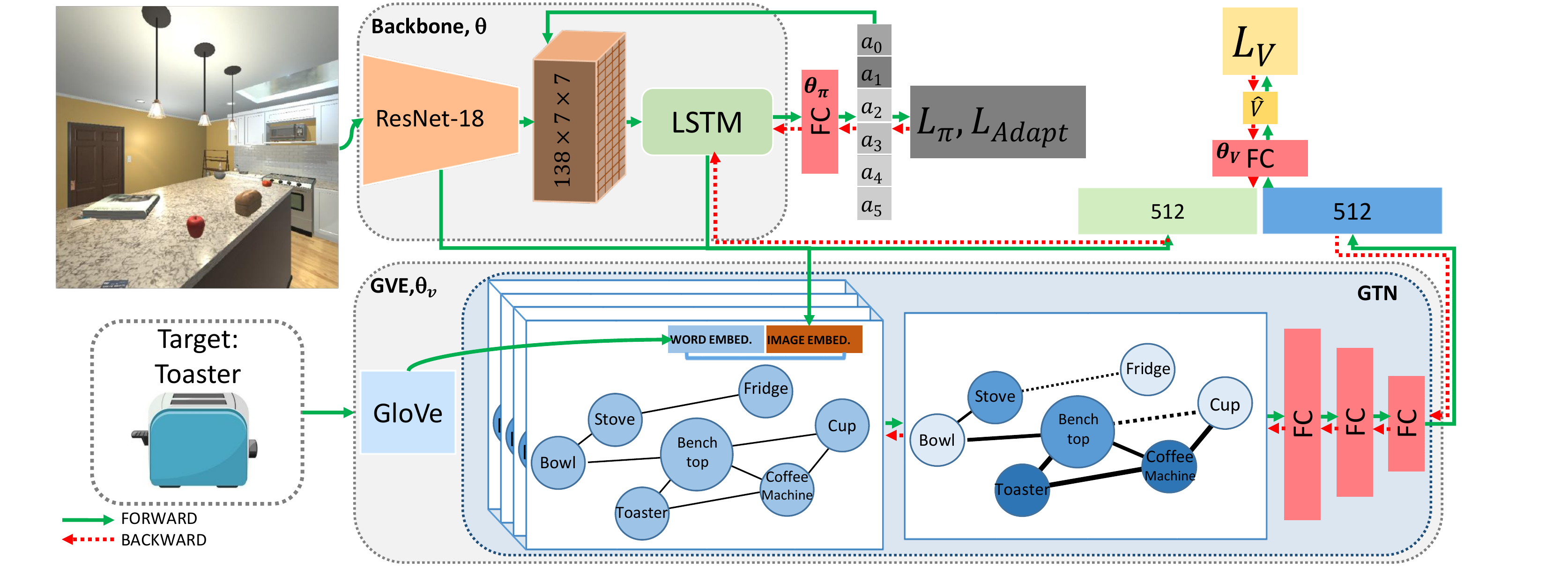}
\caption{Overview of our method. Our GVE module augments the A3C backbone and is trained using the supervised loss $L_V$ at each time step of the trajectory.}
\label{overview_fig}\vspace{-3mm}
\end{figure*} 

For the adaptation, inspired by~\cite{savn}, we adopt Model Agnostic Meta-Learning (MAML)~\cite{maml} to continuously adapt the learnt knowledge during test time. To do so, in our RL setup, we divide the training trajectories into meta-train $D$ and meta-validation $D'$ domains. $D$ includes sub-trajectories of defined intervals and $D'$ covers the complete trajectory. Then, $L_{Adapt}$, a loss function parameterised by $\bphi$ is learnt to compensate for the domain shift during test. We optimise the adaptation loss and the policy loss $L_{\pi}$ according to the following update rules:
\begin{equation}
\begin{split}
\btheta_{\text{total}}^{\textit{i}} &\xleftarrow{} \btheta_{\text{total}} - \alpha \nabla_{\btheta_{\text{total}}}L_{\textit{Adapt}}^{\tau \sim \textit{D}}(x_t, \pi(x_t); \bphi) \\
\btheta_{\text{total}} &\xleftarrow{} \btheta_{\text{total}}^{\textit{i}} - \beta_{1} \nabla_{\btheta^{\textit{i}}_{\text{total}}}L^{\tau \sim \textit{D'}}_{\pi}(x_t, \pi(x_t)) \\
\bphi &\xleftarrow{} \bphi - \beta_{2} \nabla_{\phi}L_{\pi}^{\tau \sim \textit{D'}}(x_t, \pi(x_t))
\end{split}
\end{equation}
We define $\btheta_{\text{total}}=\{\btheta, \btheta_{\pi}\}$ in the above update rules for readability, $\alpha$, $\beta_1$ and $\beta_2$ are the learning rate hyper-parameters and $\theta_{\textit{total}}^i$ is the set of intermediate weights after the adaptation before the final update. In practice we can have multiple intermediate updates $i$ before the final update.

\subsection{Prior Knowledge Graph}
Our GVE module relies on a prior knowledge graph $G(V, E)$ which contains the information about the co-location of the objects in a scene. Intuitively, if the joint probability of observing object $i$ and $j$ together in a single view is high in an environment, then observing one should provide a strong cue for finding the other. Based on this intuition, inspired by the work of \cite{scene_priors}, we define an object co-location graph. The set of nodes $V$ includes all the objects visible in the environment (\textit{e.g.} whether used for navigation target or not) represented by their features. Each node feature, $\mathbf{v}_i \in \R^{d}$, encodes the concatenation of the RGB observation features extracted using a CNN and the semantic vector embedding of the object, extracted using a word embedding model. The edges show the co-location of the objects; that is, $e_{ij}$ becomes incresingly closer to one during the training if and only if the objects $i$ and $j$ often appear in the same egocentric view of the agent, simultaneously. 

We have multiple separate graphs encoding the objects visible in different room types. This is one of the novelties of our method which is, particularly, important since similar objects might appear in different rooms while their co-locating objects might be different. This will allow for having different edge values for our graph in each room. For instance, the object "bowl" might appear both in the "kitchen" and the "living room". It might be observed near the "cabinets" in the "kitchen" but close to "sofa" in the "living room". Therefore, our adjacency matrix, $A \in \R^{n\times n\times C}$, is a three-dimensional tensor where each channel $C$ encodes the knowledge specific to a room type. We perform this by initialising the edges between objects that do not appear in each specific room type with zero. The graph separation enables the agent to mainly attend to one of the graph channels in each scene and avoid distraction. This way, more scene-specific knowledge can be encoded.

Inspired by \cite{scene_priors}, we also initialise the graph's binary edge weights using the co-occurrence of the objects in images acquired from the Visual Genome dataset \cite{visual_genome}. However, we learn more accurate edge weights along with node features during the training using the attention mechanism, as proposed in Graph Transformer Network \cite{gtn}. 

\subsection{Graph-based Value Function Estimation in Actor-Critic} \label{gve}

It is proven that policy gradient algorithms in RL have high variance of gradient estimation and different solutions have been proposed to address this \cite{var_reduction, var_reduction_2}. The issue is mainly due to the difficulty of credit assignment which is exacerbated where the reward is sparse. In actor-critic methods the variance is reduced by using the boot-strapped estimates of the state-value function as a baseline. It is also known that a more accurate estimate of the value function leads to a more optimally converged policy \cite{reinforce_1992}.

Accurate estimation of the state value in visual navigation is significantly challenging, particularly when searching for a target object. This is mainly due to the lack of global context at each time step where the agent can only observe a limited view of the scene. The context is conventionally encoded using the recurrent neural networks which are prone to forgetting and challenging to train. We propose our Graph-based Value Estimation (GVE) module to address this problem. Using GVE we encode the global context of object locations into our model so the agent has a better understanding of its position in the scene with respect to other objects. This will help the agent more realistically estimate the state value. Intuitively, there's a correlation between the objects present in each time-step's viusal input and the target object to navigate to. For example, when navigating to a smaller object like ``book'' in a room, the agent may look for larger indicator objects like desk or shelf that guide it towards the target. This means that the agent should be \textit{optimistic} about finding the target object by observing a correlated object based on its prior knowledge. In other words, a state where a correlated object is visible should still have a high estimated value, close to the value of a state where the target is visible.

In order to encode our prior knowledge graph for value estimation, we propose to use Graph Transformer Networks (GTN) \cite{gtn}. Using GTN our agent is able to learn new edges and update the edge weights. This way, depending on the scene, we are able to extract features from the graph that help with more efficient navigation. In order to update the original edge weights we first learn a soft weighted average of the edges across multiple channels of the input adjacency matrix:

\begin{equation}
    H^l_i = \text{softmax}(\boldsymbol{W^l_i})A
\end{equation}
where $W^l_i \in \R^{1\times1\times C}$ is a weight matrix over the channels. $H_i$ will be the concatenation of multiple $H^l_i$ matrices to learn multiple new mappings of the original adjacency matrix. In practice we can learn $M$ different $H_i$ and the final new adjacency matrix is then defined as the matrix multiplication of those which can encode the common paths:
\begin{equation}
    A_{\textit{NEW}} = H_1H_2 \textit{...}H_M
\end{equation}
Finally, in order to encode the node features using the new adjacency matrix $A_{\textit{NEW}}$, we use graph convolutional layers defined as:

\begin{equation}
Q = \sigma (\tilde{E}_i^{-1} \tilde{A}_{\textit{NEW}}\boldsymbol{W_N}N)
\end{equation}
where $Q$ is the final embedding of the graph, $N \in \R^{n\times d}$ is the input node features, $\boldsymbol{W_N}$ is the weights for node embedding, $\tilde{A}_{\textit{NEW}} = A_{\textit{NEW}} + I$ is the augmented learned adjacency matrix with self-connections using identity matrix $I$ and $\tilde{E}^{-1}$ is the inverse degree matrix for $\tilde{A}_{NEW}$. In practice we can learn multiple convolutional layers embedding different node features using the new adjacency matrix. Therefore, the output graph representation vector $Q$ is the result of both node and edge operations dynamically learnt during training. 

In order to incorporate the graph's encoding into our GVE, we partially separate the parameters of the critic sub-network, $\btheta_v$, from the actor sub-network $\btheta_{\pi}$. Therefore, in our method we estimate the state-value function according to:
\begin{align}
\hat{V}(x_t) &=  \boldsymbol{W_1}Q + \boldsymbol{W_2} F(x_t, Z; \btheta) \label{eq4}
\end{align}
$\boldsymbol{W_1}$ and $\boldsymbol{W_2}$ are aggregation parameters for the final value estimation which are also trained along the other parameters of the network and $\btheta$ is the set of parameters for $F$, the state encoding backbone network.

As can be seen in equation \ref{eq4}, our method decomposes the expected future reward into two components: one, estimated mainly conditioned on the state as encoded by the encoding network parameterised with $\btheta$; and the other component, $Q$, estimated by the graph encoding network which represents a correlations between the target object and the objects that are visible in the current state $x_t$. Therefore, the agent can include the expectation of the objects it might observe in the future states in order to estimate the value of current state more accurately. In other words, the agent has a more global understanding of the environment.



\section{Experiments}\label{results}
\subsection{Experimental Setup}

We choose AI2THOR environment for our experiments. This simulator consists of photo-realistic indoor environments (e.g. houses) categorised into four different room types: kitchen, bedroom, bathroom and living room. In order for fair comparison, we follow the same experimental setup as our main baseline \cite{savn}. In this setup, 20 different scenes of each room type are used for training; 5 scenes for each as validation and 5 for test. 
Following the recent conventions in visual navigation tasks, we measure the performance of our method based on Success Rate (SR) and Success weighted by Path Length (SPL).

\subsection{Implementation Details} \label{network_arch}

Our actor-critic network comprises of a LSTM with 512 hidden states and two fully-connected layers one for actor and the other for the critic. The actor outputs a 6-dimensional distribution $\pi(x_t)$ over actions using a Softmax function while the critic predicts a single scalar value. The critic sub-network also receives the GTN encoding as a vector of size 512 besides the 512-dimensional LSTM encoding. The concatenation of these two feature vectors is mapped to the value using a fully connected layer.

We use Glove~\cite{glove} to generate $300$-dimensional semantic word embedding of the objects and a pre-trained ResNet-18 to extract the features from the $300\times300$ image inputs. We concatenate these two to feed to our state encoder, LSTM. The overview of the architecture can be seen in figure~\ref{overview_fig}

Our prior knowledge graph is pre-trained on Visual Genome dataset \cite{visual_genome}. The edge weights are set to one where the objects co-occur in a scene at least three times, and zero otherwise. There are 89 nodes in each channel of the graph and 5 channels in total; 4 layers dedicated to the objects in each room type and one self-connections layer for regularisation. We train a two layer adjacency matrix using GTN. The input to the graph, as node features, is a 1024-dimensional vector. This vector is the concatenation of 512-dimensional observation features, extracted using ResNet-18, with 512-dimensional Glove~\cite{glove} embedding of the object name. The Glove embedding is mapped from 300 to 512 using a fully connected layer. The GTN features are mapped into a 512-dimensional vector using a fully connected layer.

\begin{figure}[t]
\centering
\includegraphics[width=8.5cm]{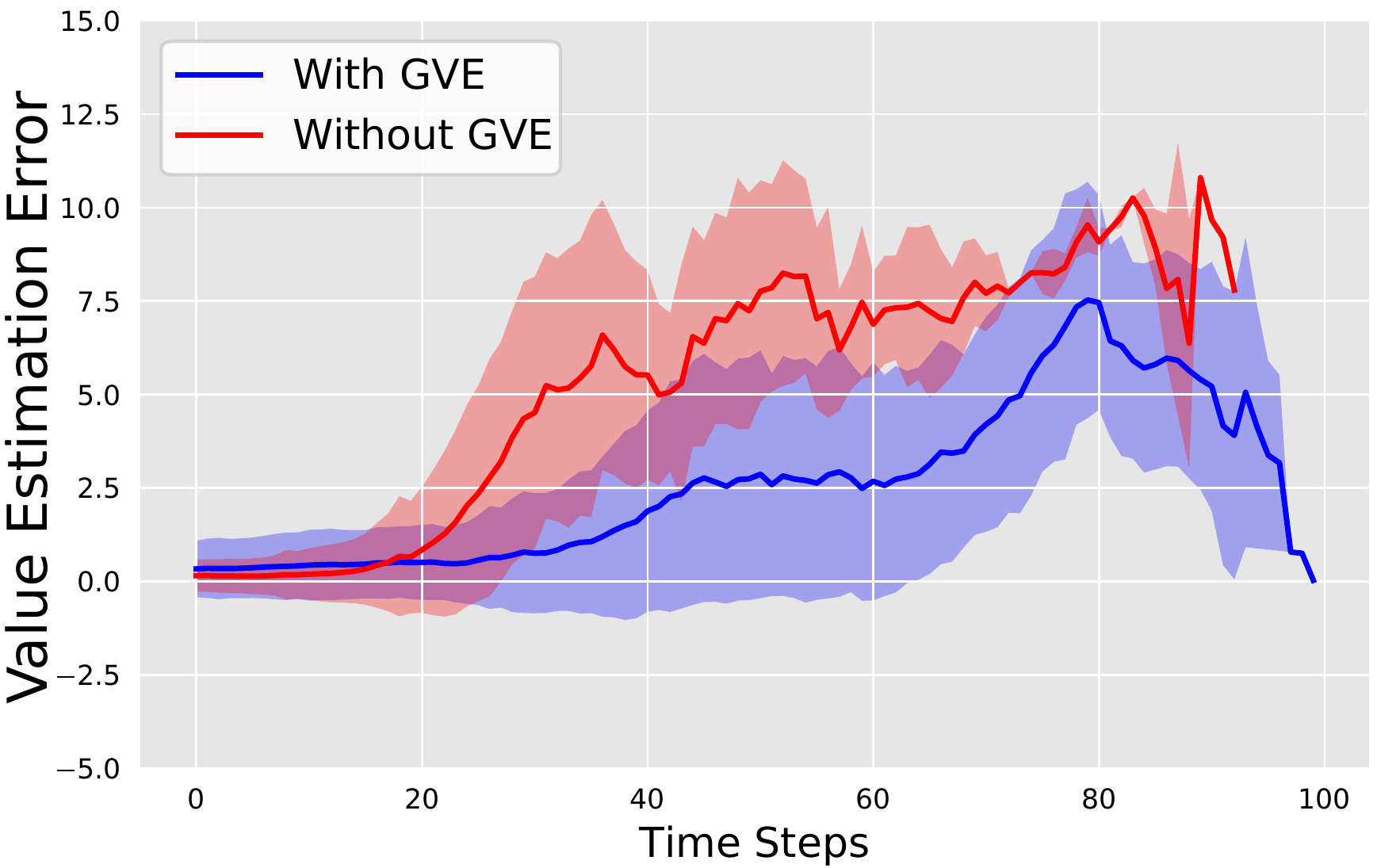}
\caption{Value estimation error on test set. Our model has more accurate estimates of the state value function (lower error) which leads to better performance.}\vspace{-3mm}
\label{gve_error}
\end{figure}

\begin{table*}[t]
\centering
\caption{Quantitative comparison of our results with the baselines. Our approach improves all the baselines in all the four evaluation metrics, conventionally used in the previous state-of-the-art methods.}
\begin{tabular}{l|l|l|l|l}
\toprule
\textbf{Method}       & \textbf{SPL} & \textbf{Success} & \textbf{SPL \textgreater 5} & \textbf{Success \textgreater 5} \\ \hline \hline
\textbf{Random}          & 3.64        & 8.01            & 0.1                       & 0.28                           \\ 
\textbf{A3C}          & 14.68 $\pm 1.8$        & 33.04 $\pm 3.5$           & 11.69 $\pm 1.9$                       & 21.44 $\pm 3.0$                           \\ 
\textbf{A3C+Graph} \cite{scene_priors}         & 15.47 $\pm 1.1$         & 35.13 $\pm 1.3$            & 11.37 $\pm 1.6$                       & 22.25 $\pm 2.7$                           \\ 
\textbf{A3C+MAML(SAVN)} \cite{savn}        & 16.15 $\pm 0.5$        & 40.86 $\pm 1.2$            & 13.91 $\pm 0.5$                       & 28.70 $\pm 1.5$                           \\ \hline
\textbf{A3C+Graph+GVE-ours}         & 16.02        & 38.22            & 13.23                       & 27.47                           \\ 
\textbf{A3C+Graph+MAML-SS-ours (w/o. GVE)}  &15.13      &38.8           & 13.68                            &29.64                                 \\
\textbf{A3C+Graph+MAML-Action-ours (w/o. GVE)} & 13.88        & 43.3             & 12.93                       & 33.53                           \\ 
\textbf{A3C+Graph+MAML+GVE-ours}  & \textbf{17.27} $\pm 0.3$ & \textbf{43.8} $\pm 1.1$    & \textbf{15.39} $\pm 0.2$                        & \textbf{33.68} $\pm 0.9$                            \\ 
\bottomrule
\end{tabular}
\vspace{-7pt}
\label{tab:results_compare}
\end{table*}

In order to train our model, we use Pytorch framework. We use SGD optimizer for meta-train weight updates and Adam~\cite{adam} for the meta-test. As for the reward, we use a constant value of 5 for reaching the target and a step penalty of -0.01 for each single step. The maximum number of steps is capped at 50 during training and 200 during testing. We train all our methods until convergence with the maximum seven million episodes, whichever occurs first.

\subsection{Baseline and SOTA Comparison}
In order to better show the contribution our method we first compare it with a few different state-of-the-art (SOTA) and baselines, shown in Table~\ref{tab:results_compare}. Firstly, we compare with the previous SOTA introduced by Wortsman~\etal~\cite{savn}, abbreviated as \textbf{A3C+MAML}. It uses MAML~\cite{maml} to learn a loss function approximated by an instance of Temporal Convolutional Networks~\cite{tcn}, over training episodes. 
Secondly, a related work \cite{scene_priors} uses a fixed knowledge graph structure to encode object relationships as part of the state space, abbreviated as \textbf{A3C+Graph}. 
In our approach, we benefit from the graph information more efficiently using our proposed GVE module. In addition to that, our graph structure and embedding architecture is also different, enabling more accurate estimation of the value function.

Furthermore, to highlight the challenging navigation scenarios that our method is tackling, we compare our results with some trivial baselines. One simple solution is a~\textbf{Random} agent for which the policy is to uniformly sample an action at all times. Another trivial baseline is shown as~\textbf{A3C} in Table~\ref{tab:results_compare}. This method is the result of dismantling our GVE and adaptation modules from the overall framework. Therefore, it acts as the simplest RL-based agent that shares the same backbone model with our proposed method.

\begin{figure*}[t]
\centering
\includegraphics[width=0.9\textwidth]{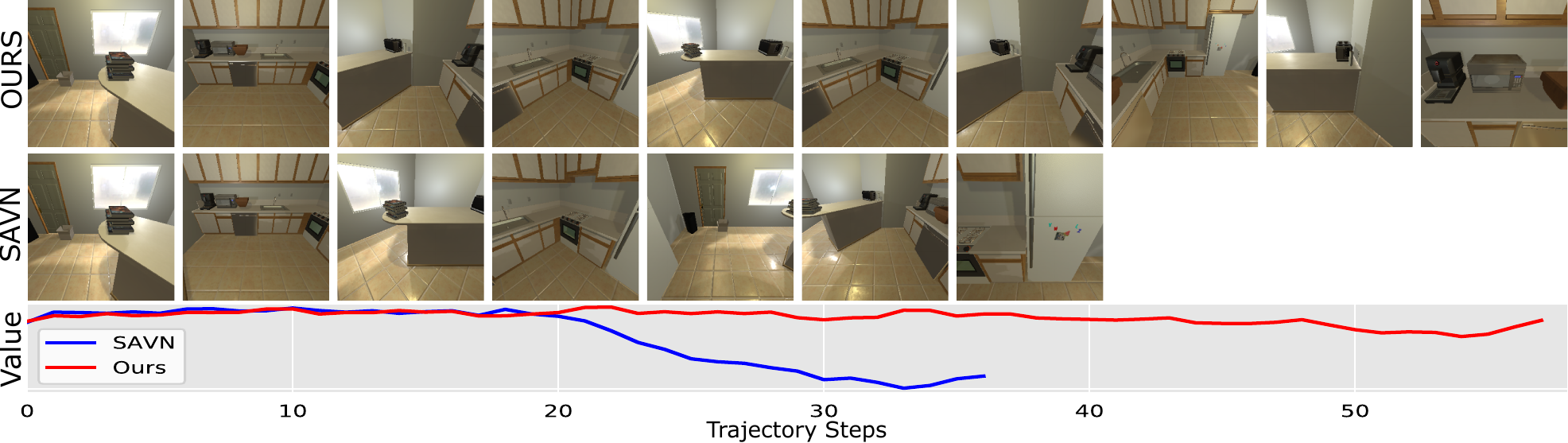}
\includegraphics[width=0.9\textwidth]{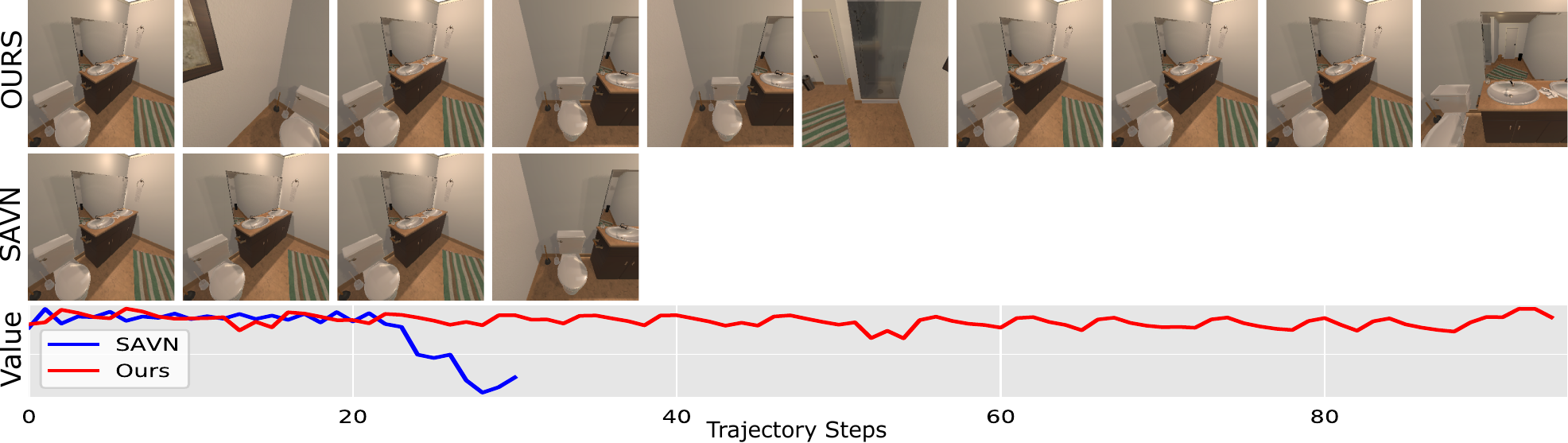}
\caption{Qualitative comparison of two sample trajectories between our method and the baseline. \textbf{Top}: target object: "bowl" in the kitchen. Our agent observes the bench-top and continuously predicts a high value (\textit{being optimistic}) until finding the target. The baseline agent, however, has a less accurate understanding of the scene and \textit{gives up} after a while not detecting the target.
\textbf{Bottom}: target object: "soap bottle" in the bathroom. Our agent observes the basin and constantly predicts a high state value even though the soap bottle is small to be detected in the visual features. The baseline gives up after a few failed actions of moving forward.
}
\label{trajectory}
\end{figure*}

\subsection{Results and Ablation Studies}
In this section we seek to answer a few principal questions with regards to our proposed method that shed more light on its strengths as well as weaknesses. 

\begin{itemize}
    \item \textit{How does GVE improve value estimation error?}
\end{itemize}

In order to respond to this question, we analyse the estimated values over the trajectories using our GVE in comparison to the baseline method. The estimation error over 1000 trajectories collected in the test scenes (unseen during training) is provided in Figure~\ref{gve_error}. The figure shows average and standard deviation of L2 distance per time step between the ground-truth value and the predictions. The red line shows the error for the baseline while the blue line shows the improvement as a results of our GVE. Our method effectively addresses the forgetting problem associated with LSTMs by estimating the values for longer trajectories more accurately.

\begin{itemize}
\item \textit{Is GVE the optimal way to incorporate prior knowledge?}
\end{itemize}

In order to show the effectiveness of our graph-based value estimation, 
we compare our final method with two variants. First variant, termed as \textbf{A3C+Graph+MAML-SS} is to simply add our graph as part of the state space observation to the backbone network. As can be seen in Table~\ref{tab:results_compare} this method improves SR but not the SPL compared to \textbf{A3C+Graph}; however, compared to \textbf{A3C+MAML} in degrades the performance. We conjugate that the reason is by adding the complex graph to the backbone network, the size of the state space is increased by orders of magnitude which impedes learning an effective state encoding under sparse reward constraint.

In another variant, termed as \textbf{A3C+Graph+MAML-Action}, we directly incorporate the graph into the policy for better decision making. Therefore, the policy will take actions conditioned upon the graph encoding. This approach can be observed as a weighted ensemble of policies representing different distributions. This leads to an improvement on SR but a significant degradation of SPL. We hypothesise that this can help the policy on better detecting the target object and thus on-time stopping which increases the SR; however, it cannot help with the optimality of the final policy. This further proves the effectiveness of our proposed GVE.

\begin{table}
\vspace{-3pt}
\caption{\emph{Ours-LG} is the variant of our model where the image features are removed from the graph node features. We can see the graph is highly reliant on the observations to learn the relationships.}
\vspace{1pt}
\centering
{\small
\begin{tabular}{l|c|c}
\toprule
\textbf{Method}      & \textbf{SPL} & \textbf{Success} \\ \hline \hline
\textbf{Ours-LG}  & 12.84         & 42.4           \\ 
\textbf{Ours-GVE} & 17.27       & 43.8   \\ \hline
\end{tabular}}
\vspace{-5pt}
\label{tab:graph_node}
\end{table}

\begin{itemize}
    \item {What is the contribution of prior knowledge for value estimation?}
\end{itemize}

In order to show the significant contribution of prior knowledge in our current framework we train our final model with randomly initialised edge weights in the graph. As can be seen in \ref{tab:randomgraph}, adding the graph with random edge weights does not improve the results compared to our baseline methods. This further confirms that our GVE module is reliant on the prior knowledge for more accurate value estimation.

\begin{table}
\vspace{-3pt}
\caption{\emph{Ours-RandomGraph} is the variant of our model where the edge weights are initialised randomly. The results on this table show the contribution of prior knowledge in our framework.}
\vspace{2pt}
\centering
{\small
\begin{tabular}{l|c|c}
\toprule
\textbf{Method}      & \textbf{SPL} & \textbf{Success} \\ \hline \hline
\textbf{Ours-RandomGraph}  & 15.32         & 40.5           \\ 
\textbf{Ours-GVE} & 17.27       & 43.8   \\ \hline
\end{tabular}}
\vspace{-5pt}
\label{tab:randomgraph}
\end{table}

\begin{itemize}
    \item \textit{What is the quantitative improvement with respect to the baselines and previous state-of-the-art methods?}
\end{itemize}
As can be seen in Table~\ref{tab:results_compare}, our final model \textbf{A3C+Graph+MAML+GVE} improves the previous SOTA~\textbf{A3C+MAML} by almost 3\% on SR and more than 1\% on SPL. This shows that our GVE helps the agent to find more targets in smaller number of steps, as a result of having a more optimal policy. Our method is, particularly, more effective on longer trajectories where it improves the baseline by almost 5\% on SR and almost 2\% on SPL. This, again, confirms the effectiveness of our GVE which can provide enough global context disentangled from the length of the trajectory.



\begin{itemize}
    \item \textit{Is providing visual features to the graph nodes helpful?}
\end{itemize}

In order to show the integrity of current design, we show the effect of removing observation (egocentric image) features from the node feature matrix. Thus, the graph will reduce to a fixed correlations among the objects, given in language embedding only (hence the name Language Graph, LG). It can also be observed as a sub-network for the value estimation to store value decomposition information without considering the observational correlations.
As can be seen in Table~\ref{tab:graph_node}, the performance of the model \textbf{Ours-LG} is significantly lower, particularly in SPL. This shows that the graph's contribution is not due to simply adding more trainable parameters.

\begin{itemize}
    \item \textit{How is our more optimal policy performing qualitatively?}
\end{itemize}{}

Finally, we provide qualitative comparison of the performance of our method compared to previous SOTA~, \textit{e.g.} \textbf{A3C+MAML}~\cite{savn}. As is shown in Figure~\ref{trajectory} the agent is navigating towards an instance of "microwave" in a kitchen scene. Our agent is able to find the target while constantly predicting a high value for the states due to observing the related objects like the "benchtop". The baseline, however, is after a few steps predicts low state values and \textit{gives up} without reaching to the target.

For more detailed comparison, we also provide performance results per room type, in Table~\ref{tab:per_room}. Additional trajectory visualisations can be found in supplementary material.

\begin{table}[]
\centering
\caption{Detailed comparison with the baseline method; SPL/Success rate are reported per room type. We can see that our method is general enough that improves the performance in 3/4 of the room types, with marginal performance on 1/4.}
\scalebox{0.8}{
\begin{tabular}{l|c|c|c|c}
\toprule
\textbf{Method}      & \textbf{Bathroom} & \textbf{Bedroom} & \textbf{Kitchen} & \textbf{Living} \\ \hline \hline
\textbf{A3C+MAML}  & 28.49/69.6      & \textbf{8.65/29.2}    & 17.8/43.6     & 7.71/21.6            \\ \hline
\textbf{Ours-GVE}  & \textbf{31.03/75.6}     &8.06/27.6      & \textbf{17.93/45.6}     & \textbf{9.41/25.2}  \\ \hline
\end{tabular}
}\vspace{-5mm}
\label{tab:per_room}
\end{table}

\section{Conclusion}
In this paper we present a simple yet effective method to improve the performance of the actor-critic RL algorithm for visual navigation. Our method improves the state value function estimation accuracy by incorporating object co-location information in the form of a neural graph. Actor-critic RL is heavily reliant on the accuracy of the state value estimation to converge to the optimal policy. This is mainly because accurate value estimation helps with correct identification of the actions along the trajectory that contribute the most to the final reward. Through extensive empirical studies we show that our agent is realistically more optimistic (\textit{e.g.} accurately predicts higher state values). This leads to successful navigation towards the target object when the baseline agents usually give up.

{\small
\bibliographystyle{ieee_fullname}
\bibliography{main_paper}
}

\end{document}